\documentclass[conference]{IEEEtran}
\IEEEoverridecommandlockouts
\usepackage{cite}
\usepackage{amsmath,amssymb,amsfonts}
\usepackage{algorithmic}
\usepackage{graphicx}
\usepackage{textcomp}
\usepackage{xcolor}
\usepackage{hyperref}
\def\BibTeX{{\rm B\kern-.05em{\sc i\kern-.025em b}\kern-.08em
    T\kern-.1667em\lower.7ex\hbox{E}\kern-.125emX}}
\begin{document}

\title{Butterflies : A new source of inspiration for futuristic aerial robotics
}
\author{Chakravarthi Jada, Lokesh Ch.R.S, Ashok Urlana, Shridi Swamy Yerubandi , \\Kantha Rao Bora, Gouse Basha Shaik, Pavan Baswani, Balaraju Karri \\ 
 RGUKT-Nuzvid, India \\
 cv@rguktn.ac.in}


\maketitle




\section{INTRODUCTION}

Nature is an inhabitant for enormous number of
species. All the species do perform complex activities with
simple and elegant rules for their survival. The property of
emergence of collective behavior is remarkably supporting
their activities. One form of the collective behaviour is the
swarm intelligence - all agents poses same rules and capabilities. This equality along with local cooperation in the agents
tremendously leads to achieving global results. Some of the
swarm behaviours in the nature includes birds formations
, fish school maneuverings, ants movement. Recently, one
school of research has studied these behaviours and proposed
artificial paradigms such as Particle Swarm Optimization
(PSO), Ant Colony Optimization (ACO), Glowworm Swarm
Optimization (GSO) etc, \cite{b1}\cite{b4}. Another school of research
used these models and designed robotic platforms to detect
(locate) multiple signal sources such as light, fire, plume,
odour etc. Kinbots platform \cite{b4} is one such recent exper-
iment. In the same line of thought, this extended abstract
presents the recently proposed butterfly inspired metaphor
and corresponding simulations, ongoing experiments with
outcomes.

\section{BUTTERFLIES AS NEW INSPIRATION}

\begin{figure}[ht]
    \centering
    \includegraphics[width=3.6in]{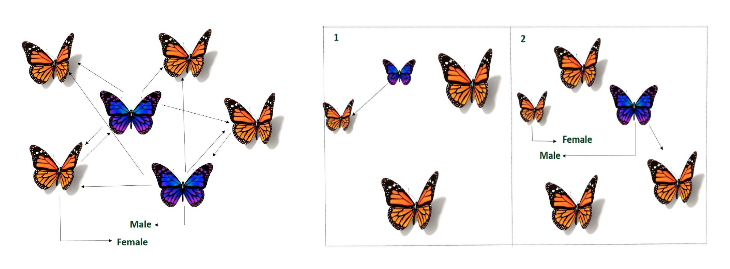}
    \caption{Pictorial representation (a). Patrolling (b). Perching}
    \label{fig:fig1}
\end{figure}

The close watch of bizarre flying of the butterflies in
the nature leads to many queries. The state of art indicated
that, butterflies mainly does the communication for the
sake of mating. Butterflies mates based on either patrolling
(Fig \ref{fig:fig1}.a) - where the female butterflies reflect UV to male
butterflies based on the distances, and the male will respond,
or perching (Fig \ref{fig:fig1}.b) - where the male sit at the hill top and
search in its territory among females passing by, and further
they use other traits such as size and fluttering to select the
mate.
By considering the above mentioned behaviours and vir-
tual simulations done by sowmya et. al \cite{b3}, chakravarthi
et. al \cite{b2} proposed \textbf{Butterfly Mating Optimization} (BMO) for simultaneously capturing all local optima of multimodal
functions. BMO basically uses patrolling mating behaviour.
BMO algorithm does not distinguish between male and
female. Hence the ``butterfly in nature named as Bfly in
the search space”. The four phases of the algorithm are: (a).
UV updation: each Bfly updates its UV value in proportion
to the fitness (signal strength) value at the present location of
Bfly. (b). UV distribution: each Bfly distribute its UV to the
remaining Bflies according to the distance from its location.
(c). l-mate selection: each Bfly select its l-mate (local) as
per the UV and fitness values \cite{b2}. (d). Movement: each Bfly
moves towards its l-mate to a step-size specified. See \cite{b2} for
more understanding of the mathematics of the phases.

\section{SIMULATIONS AND EXPERIMENTS}
\begin{figure}[ht]
    \centering
    \includegraphics[width=3.5in]{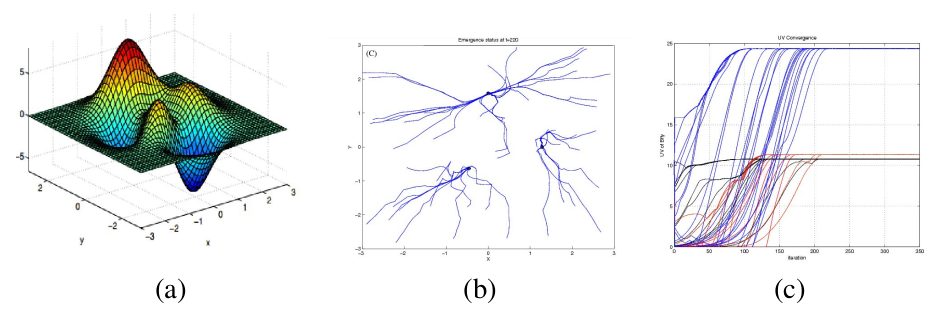}
    \caption{(a). Function (b). Emergence (c). UV convergence}
    \label{fig:fig2}
\end{figure}

\begin{figure}[ht]
    \centering
    \includegraphics[width=3.5in]{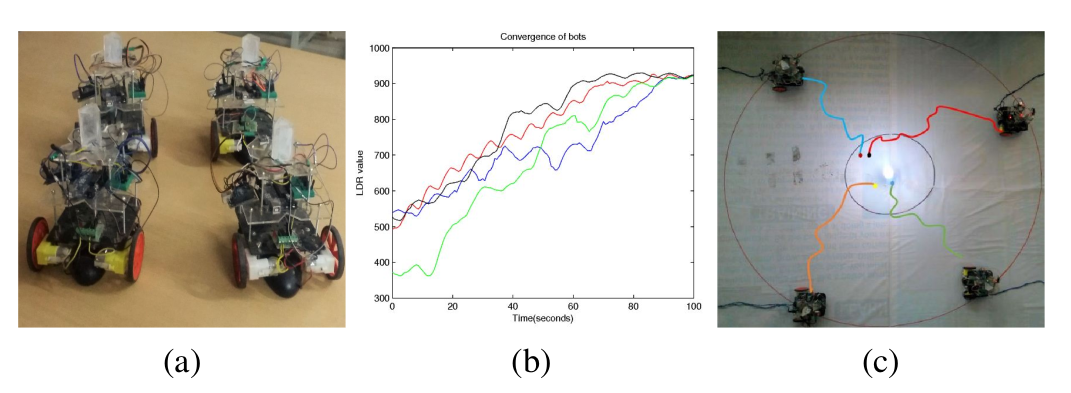}
    \caption{(a). BflyBot swarm (b). Luminescence convergence
(c). Bots emergence}
    \label{fig:fig3}
\end{figure}

The BMO algorithm has formulated to capture the local
optima of multimodal functions. To check the efficacy of the
algorithm, it has been applied to 3-D multimodal function.
Fig \ref{fig:fig2}.a, b, c shows the 3-D function, emergence of all Bflies,
UV convergence for all the iterations. Though the BMO
shares some features with GSO algorithm, it has got its own
distinctive qualities, and results are quite impressive in terms
of smoothness in the emergence paths and l-mate selection
variation. By looking at the efficacy of the BMO, we built
mobile robotic swarm - BflyBots, which mimic the Bflies in
the BMO. Various experiments are conducted to detect the
single fluorescent signal source. Fig \ref{fig:fig3}.a shows the swarm of
four mobile BflyBots. Fig \ref{fig:fig3}.b, c shows the emergence paths
of all bots and fluorescence convergence for all time steps.

\subsection{RESULTS AND DISCUSSION}\label{AA}
The experimental results indicating the efficacy of the
BMO in detecting the light source. We checked it with
varying initial locations of the four bots. The experiments
also conducted by varying step-size in the process of de-
tection, and found that smoothness and time of co-location
are effecting. Currently, we are increasing the number of
bots and looking forward to conduct experiments to see the efficacy of the paradigm in deteting multiple light sources.
In the same line of thought, we also planned to modify the
algorithm for the sake of capturing dynamically dislocating
light sources. We also planned to take a step to make our
trails to embedded the model into multi-quad rotor swarm,
which could suits to many practical applications.

\end{document}